# Can Grammarly and ChatGPT accelerate language change? AI-powered technologies and their impact on the English language: wordiness vs. conciseness

*¿Pueden Grammarly y ChatGPT acelerar el cambio lingüístico? Tecnologías basadas en IA y su impacto en la lengua inglesa: palabrería frente a concisión*

**Karolina Rudnicka**
Faculty of Languages, University of Gdańsk, Poland
karolina.rudnicka@ug.edu.pl

**Abstract:** The proliferation of NLP-powered language technologies, AI-based natural language generation models, and English as a mainstream means of communication among both native and non-native speakers make the output of AI-powered tools especially intriguing to linguists. This paper investigates how Grammarly and ChatGPT affect the English language regarding wordiness vs. conciseness. A case study focusing on the purpose subordinator *in order to* is presented to illustrate the way in which Grammarly and ChatGPT recommend shorter grammatical structures instead of longer and more elaborate ones. Although the analysed sentences were produced by native speakers, are perfectly correct, and were extracted from a language corpus of contemporary English, both Grammarly and ChatGPT suggest more conciseness and less verbosity, even for relatively short sentences. The present article argues that technologies such as Grammarly not only mirror language change but also have the potential to facilitate or accelerate it.

**Resumen:** La proliferación de tecnologías lingüísticas basadas en la PLN, los modelos de generación de lenguaje natural basados en la IA y el hecho de que el inglés sea un medio de comunicación mayoritario entre hablantes nativos y no nativos hacen que el resultado de estas herramientas resulte especialmente intrigante para los lingüistas. Este artículo investiga cómo Grammarly y ChatGPT están afectando a la lengua inglesa en lo que respecta a la concisión frente a la prolijidad. Se presenta un estudio de caso centrado en el subordinador de propósito *in order to* ilustrar cómo Grammarly y ChatGPT recomiendan estructuras gramaticales más cortas a cambio de otras más largas y elaboradas. Aunque las frases analizadas fueron producidas por hablantes nativos, son perfectamente correctas y se extrajeron de un corpus lingüístico de inglés contemporáneo, tanto Grammarly como ChatGPT sugieren más concisión y menos verbosidad, incluso para frases relativamente cortas. La presente comunicación sostiene que tecnologías como Grammarly no solo reflejan el cambio lingüístico, sino que también tienen el potencial de facilitarlo o acelerarlo.
**Palabras clave:** Grammarly, ChatGPT, cambio lingüístico, longitud de las frases.

# 1 Introduction

Any language, spoken or written, changes over time. Both vocabulary, grammatical rules, and, in particular, the subjective linguistic norms that define what we consider correct and what we do not are evolving. It is common knowledge that English, which is nowadays the universal tool for communication in a globalised world community, has changed quite dynamically throughout its history, both in terms of grammar and lexis (i.a. Akimoto et al., 2010; Durkin, 2014; Hickey, 2012; Mair, 2006).

The impact of language change is most visible in structure and form on the constructional level, e.g. when it comes to new constructions rising in frequency or loanwords entering the language. However, a significant role is played by higher-order changes or processes which operate in the background and influence whole constructional networks and larger organisational units of the language (Hilpert, 2013: 14; Kempf, 2021; Rudnicka, 2019, 2021a,b,c). In the literature, such changes are also sometimes referred to as system-dependency (Hiltunen, 1983; Hundt, 2014; Petré, 2010). Among relatively recent developments which we can see as externally-motivated[1] higher-order processes, there are the socio-cultural changes of the late nineteenth and early twentieth centuries, instantiated by the invention of new printing technologies, the development of mass literacy, the advent of the mass-circulation press, and the invention of the telegraph (described by Hames and Rae, 1996). These processes led to changes in the structure of the English language, such as the shortening of sentence length in terms of words (Fries, 2010; Gross et al., 2002; Rudnicka, 2018, 2019); the adoption of new punctuation conventions (Fahnestock, 2011; Rudnicka, 2018); the gradual decline of purpose subordinators, such as *in order to*, *in order that* or *so as to*, from the core grammar of the English language (Rudnicka, 2019, 2021) and their replacement by shorter equivalents such as the *to*-infinitive (Los, 2005). Another example of the fundamental role played by higher-order changes is the loss of German *so*-relatives described by Kempf (2021). In her work, the author claims that the change she investigates was brought about by a hierarchy of different factors, conditioned (Kempf, 2021: 319) "to a high degree by socio-cultural and socio-linguistic developments, such as democratization, the Enlightenment and literalization."

Also, even if it sounds like a cliché, the world and the reality around us also constantly change. Much was said about the impact of short forms of communication, such as email or text messages, on language, language learning, and communication (i.a. Geertsema et al., 2011; Filipan-Žignić et al., 2016; Mehrabi and Bataghva, 2016; Tagg, 2015). In contrast, so far[2], not much attention has been paid to the overall impact on the English language of AI-powered (AI - artificial intelligence, hereafter AI) language technologies using natural language processing (hereafter NLP) such as Grammarly or ChatGPT. Since the popularity of such technologies is increasing dramatically[3,4,5], the investigation on whether and how they can influence the language seems to be a worthwhile task, which, according to the Author's best knowledge, has not been dealt with before. The present paper contributes to filling this gap and looks at the possible role of AI-based technologies in the processes of language change. It considers the possibility that these tools not only mirror but can also accelerate language change.

The article is organized in the following manner: Section 2 contains a short presentation of Grammarly and ChatGPT, whereas Section 3 deals with the methods, objectives, and case studies to illustrate the arguments and hypotheses. Finally, Section 4 discusses the findings, offers concluding remarks, and presents the study's outlook.

---

[1] *Externally-motivated* means that the change comes from outside of the language – the terminology stems from Hickey's work (2012: 42), according to whom changes which are triggered by social factors can be seen as externally-motivated changes.

[2] Until now, most research on AI-based language technologies has focused on English language teaching and how learners process what they are taught (e.g. O'Neill and Russel, 2019; Barrot, 2020).

[3] According to Grammarly's official webpage, the number of users increased from "1 million daily active users in 2015 to 30 million in 2020."
(https://www.grammarly.com/blog/grammarly-12-year-history/, accessed on March 6, 2023).

[4] Grammarly was recognized by TIME as one of the 100 most influential companies in 2022.
(https://time.com/collection/time100-companies-2022/6159466/grammarly/, accessed on March 6, 2023).

[5] According to Reuters (2023), ChatGPT is estimated "to have reached 100 million monthly active users just two months after launch, making it the fastest-growing consumer application in history (…)."
(https://www.reuters.com/technology/chatgpts-popularity-explodes-us-lawmakers-take-an-interest-2023-02-13/, accessed on March 10th, 2023).

## 2 The two AI-powered technologies in this study

As AI/NLP-based writing assistants and content generators become more commonplace, with over a hundred tools available on the Internet at present – Copysmith, Sudowrite, Trinka, ProWritingAid, Wordtune, and Instatext to name just a few – this paper zeroes in on two of the most well-known and heavily used technologies: Grammarly (a writing assistant) and ChatGPT (which generates new content). The decision to focus on these two was based on their high usage levels and accessibility. The basic versions are open for use free of charge.

### 2.1 Grammarly

Let us start with some facts concerning Grammarly. Launched in 2009 by three Ukrainian-born entrepreneur engineers[6], Grammarly is a cloud-based AI-powered writing assistance technology. Initially, it was mainly used for correcting grammar mistakes and assisting students in honing their writing skills. However, over the past ten years, the company has worked on creating more sophisticated and complex feedback systems. According to the official website, Grammarly offers a user-friendly experience that can, in real-time, detect errors in spelling, syntax, grammar, and punctuation in texts written in English (both in the free basic version and the premium and business versions). While working with Grammarly, users can personalize their writing style and detect plagiarism (only available in the premium and business versions). As previously noted, as of 2023, around 30 million people and 50,000[7] teams worldwide utilize Grammarly daily; these numbers are rising, making it one of the most popular and praised grammar correction tools and writing assistants[8] in today's world, employed by both native and non-native speakers of English[9].

### 2.2 ChatGPT

ChatGPT is an AI chatbot which interacts with its users in a conversational way. It was developed by OpenAI and launched in November 2022. As stated in Reuters (2023, cited in footnote[5] above), according to estimations, ChatGPT reached 100 million monthly active users just two months after launch, which makes it the "fastest-growing consumer application in history." In spite of the fact that it is a new tool, scientists already use ChatGPT extensively and produce research papers about it (i. a. Gordijn and Have, 2023; Van Dis et al, 2023; Doshi et al., 2023; Looi, 2023). Furthermore, according to Looi (2023), the chatbot has already appeared as a co-author on multiple research papers.

ChatGPT is very different from Grammarly, because the latter does not produce content – its main aim being to correct/enhance the way in which texts are written. However, ChatGPT is also able to help us in this regard. Since the first question one gets from it usually is "Hello! How can I assist you today?" – we can ask ChatGPT to correct or enhance a text that we prepared earlier, and ChatGPT is definitely able to assist us with this task.

On the other hand, we expect that the algorithm of ChatGPT does not differentiate between creating new content or modifying existing text. All in all, what matters are tailored probabilistic models which match words and constructions in an "optimal" sequence. Therefore, for the sake of the current study, we hypothesize that tendencies observed for enhanced texts will also apply to its other uses.

## 3 Study variable: wordiness vs. conciseness

The case study variable chosen for the present work is the trend toward shorter sentences and more concise content. The present study refers to it as *wordiness vs. conciseness*. It was chosen due to the fact that it is well-documented, with works by Fries (2010), Gross et al. (2002), Lewis (1894), Westin (2002), Biber and Conrad (2009), Rudnicka (2018, 2019), all showing that the length of a sentence written in English has been decreasing during the last two to four hundred years (Lewis, 1894). Furthermore, the decrease appears to be especially dramatic since the beginning of the twentieth century (Rudnicka, 2018, 2019). For instance, in the COHA genre *magazine*, sentences at the onset of the twenty-first century are approximately ten words shorter than at the beginning of the nineteenth century. For *non-fiction* the difference amounts to 8 words. Texts in *newspaper* genre are almost 5 words shorter now than they used to be in the middle of the nineteenth century.

---

[6] Max Lytvyn, Alex Shevchenko and Dmytro Líder, https://www.grammarly.com/about, accessed on March 6, 2023.
[7] See Grammarly's oficial website - https://www.grammarly.com/about, accessed on March 6, 2023.
[8] According to many different online-based rankings, for example https://digital.com/best-grammar-checker/, accessed on March 6, 2023.

[9] According to Grammarly's survey conducted in winter 2011, 12, 68% of Grammarly's users were native speakers of English, compared to 32% non-native speakers.

Among the possible reasons for the acceleration of this decrease, there are the major socio-cultural changes of the nineteenth and twentieth centuries already listed in the introduction, i.a., the development of mass literacy and readership, the invention of the printing press, and the competition for the reader on the newspaper market, and the change of punctuation conventions which, too, might have been brought about by the development of mass readership.

The present work applies Grammarly and ChatGPT to show how AI-powered technologies reflect the trend toward more concise content, and are potentially able to further boost it. The paper argues that given enough users and popularity, the advent of tools such as Grammarly and ChatGPT might furtherly accelerate language change processes, here exemplified by i) the decrease in the frequency of use of more elaborate phrases which have shorter equivalents, such as *in order to*; and, partly resulting from i), ii) the decrease in overall sentence length.

## 3.1 The empirical part

In the empirical part of this paper, we analyse Grammarly's and ChatGPT's output to find out if and how the technology favours and encourages writing that is more concise and to the point, with shorter sentences and simpler grammatical constructions.

Additionally, we want to see if using the technology to "correct," "enhance," or "modify" our writing would result in changes in readability. To accomplish this task, we designed a case study consisting of two parts:
1) In the first part (Sections 3.1.1 and 3.1.2) we focus on the question if Grammarly and ChatGPT judge the purpose subordinator *in order to* as redundant and unnecessary in various sentences extracted from the Corpus of Contemporary American English (COCA);
2) In the second part (Sections 3.1.3 and 3.1.4), we test the sentences constituting the output of Grammarly and ChatGPT to compare the sentence length and readability of the pre- and post-modified versions.

For both parts, we work on one hundred random sentences extracted from the COCA corpus, which belongs to the largest publicly available corpora of the English language. It covers the period from 1990 and 2019. The language variety represented in the corpus is modern American English. The COCA corpus contains one billion words in nearly half a million of different texts, which belong to eight genres, namely: *newspapers*, *popular magazines*, *fiction*, *spoken*, *academic journals*, *blog*, *web pages*, *tv and movie subtitles*.

So, the first step is the extraction of the test sentences from the corpus. We use the interactive online interface. Let us start from the first part, focused on *in order to*, before we move on to the second part of the study.

### 3.1.1  *In order to*: methodology

*In order to* is one of the most commonly used non-finite purpose subordinators in modern English. According to OED Online[10], its earliest recorded use was back in 1609.

Starting from the beginning of the twentieth century onwards, its usage decreased; it had its highest frequency per million words in the Corpus of Historical American English (COHA) – 95.72 – during the 1910s, and its lowest frequency per million words – 52.32 – in the 2000s. However, this decrease in the frequency of use is not nearly as drastic as in the case of the finite purpose subordinator *in order that*. The frequency per million words of *in order to* in COHA was 93.73 in 1900 and 52.32 in 2000. In contrast, for *in order that*, the respective frequencies of use per million words are 52.32 and 0.68 (Rudnicka, 2019: 64).

---

[10] Oxford Dictionary Online, s.v. *in order to*, retrieved on March 16, 2023 from from http://www.oed.com.

A shorter and nearly synonymous alternative for *in order to* is, according to Fowler's Dictionary of Modern English Usage (2015: 416), the purposive *to*-infinitive. Greater formality of *in order to* is suggested as one of the potential reasons for the choice of this form over the *to*-infinitive. Why would language users pick *in order to* and not simply a *to*-infinitive? Both Fowler's Dictionary and Schmidtke-Bode (2009: 174) offer a similar explanation, as they list the presence of a *to*-infinitive in the immediate proximity as one of the reasons for the choice of *in order to*. Schmidtke-Bode (2009) and Los (2009) suggest that the usage of the *to*-infinitive extended considerably in the past, and the addition of *in order* in front of *to* was an answer to that. So, the English language users might choose *in order to* over *to*-infinitive for the purposes of clarity, unambiguousness, or the need for more formality.

Summarizing, even if declining in the frequency of use, *in order to*, without doubt, as of 2023, still constitutes part of the core grammar of the English language. It is grammatical, acceptable, and present in language corpora such as COCA and COHA.

Interestingly, however, AI-based technologies such as Grammarly find more elaborate and complex phrases like *in order to*, *so as to*, *in spite of the fact that* redundant and superfluous, as even basic exploration of these tools shows. This study aims to find out if two tech products advise users to remove *in order* before an infinitive when seeing a random sample of 100 sentences featuring the purpose subordinator, taken from COCA.

Since we aim to look at a dataset that contains sentences of different lengths and represents writing typical of multiple registers, we do not only focus on one genre, so that the data set is less homogenous. The hypothesis is that at least some of the sentences will have *in order* flagged as too verbose. The research question is how many and what patterns can be noticed. To answer this research question, the following steps are taken:

1) We create a first sample of data, randomly extracting one hundred instances of sentences containing *in order to* from the COCA corpus.

2) The above sentences are provided to Grammarly's and ChatGPT's online interfaces respectively.

3) In Grammarly, the "target" domain and "audience" are set to default ("general") as is the formality ("neutral"), the options we have in the free version. In ChatGPT, we ask the chatbot if it "could rewrite the sentences so that they are better." We do not elaborate on how we define "better," we simply ask this question followed by a list of sentences.

4) We obtain the output from the AI-based technologies. At the end, we compare raw frequencies of *in order to* from the three different samples (original texts, output from Grammarly, output from ChatGPT).

### 3.1.2 *In order to*: results

Both Grammarly and ChatGPT process the sentences in real time. In Grammarly, we obtain suggestions in the form of underscores of different colours, see Fig. 1. Only after we click on a given underscore, do we get a suggestion, see Fig. 2. For the purpose of the present study, all the suggestions and modifications of the technology are accepted. In the case of ChatGPT, the chatbot just provided us with a list of rewritten sentences.

Table 1 below contains the results showing how many instances of *in order to* stayed in the three datasets.

| The dataset | Original set of one hundred sentences from COCA | Sentences processed with Grammarly | Sentences processed with ChatGPT |
| --- | --- | --- | --- |
| Raw frequency of *in order to* | 100 | 0 | 5 |

Table 1: The raw frequency of *in order to* in the three datasets.

As we can see, both Grammarly and ChatGPT greatly reduced the number of *in order to* in the sentences. In the case of the former, all of the instances disappeared from the dataset, while for the latter, as much as 95% of the cases disappeared.

The results suggest that the two AI-powered technologies show a strong "dispreference" for more elaborate purpose subordinators such as *in order to*. When it comes to Grammarly, the explanation for the suggestion to remove it, is the fact that the phrase "may be wordy" (see Fig. 2). In the case of ChatGPT, we do not exactly know why almost all the cases disappeared, since we only wanted the chatbot to "rewrite the sentences so that they are better."

In order to enjoy life, we should not enjoy it too much.

Fig. 1: Grammarly suggesting we should consider changing the underlined construction.

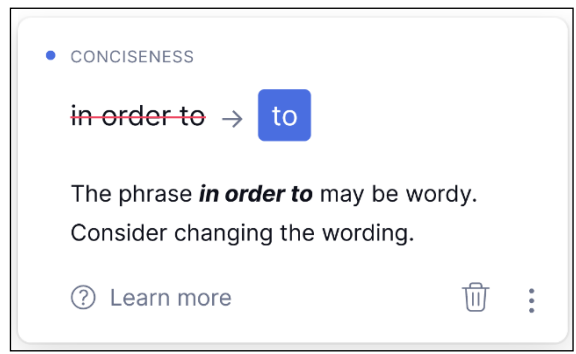

Fig. 2: Grammarly's suggestion for the sentence from Fig. 1.

### 3.1.3 Sentence length and textual complexity: methodology

In the present part, sentence length and readability of the three sets of sentences are compared. Sentence length can be, very intuitively, defined as the number of words from the first, capitalised word, to the end punctuation sign, such as the period, a question mark or an exclamation mark.

On the other hand, textual complexity is a metric that describes the difficulty of a text for its reader (Naderi et al., 2019). There are various tools and methods available to quantify textual complexity, such as readability measures (or readability formulas and indices) like the Automated Readability Index, Gunning Fog index, Flesch Kincaid Grade level, and the Flesch reading-ease test.

These measures provide quantitative scores for assessing and rating the difficulty of any given text (e.g. Fahnestock 2011: 170). They can be used to customize materials for specific audiences including schoolchildren, pupils, students and language learners, or tweak an existing text to meet certain readers' needs (Jin and Lu, 2018).

One of the most popular readability measures is the Flesch reading-ease test which measures text complexity and is only applicable for adult material at fourth grade level or higher. It takes both sentence and word length into account when adjusting content for a particular target audience. The present paper applies the Flesch reading-ease test to quantify the estimated difficulty of the test sentences before and after the sentences are modified by Grammarly and ChatGPT.

In order to obtain the scores for each dataset, an online Text Readability calculator is applied:https://readabilityformulas.com/freetests/six-readability-formulas.php#. The three sets of sentences are pasted one after the other into the online interface. The interpretation of readability scores is conducted with the use of values from Table 2 (Klare, 1975: 236)

| Score | Meaning |
|---|---|
| 100–90 | Very easy to read |
| 90–80 | Easy to read |
| 80–70 | Fairly easy to read |
| 70–60 | Standard |
| 60–50 | Fairly difficult to read |
| 50–30 | Difficult to read |
| 30–0 | Very difficult to read |

Table 2: Interpretation of the readability scores.

### 3.1.4 Sentence length and textual complexity: results

The two histograms compare the lengths of sentences from the original dataset (COCA), with the processed sentences (Grammarly in Fig. 3 and ChatGPT in Fig. 4, respectively). We observe a visible shift towards shorter sentences in both cases in the processed samples. In particular, all longer "outlier" sentences appear to become substantially shorter.

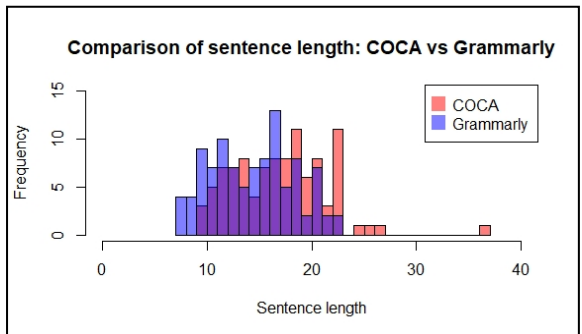

Fig. 3: Histograms presenting sentence lengths in the original dataset (COCA) and in the dataset processed by Grammarly.

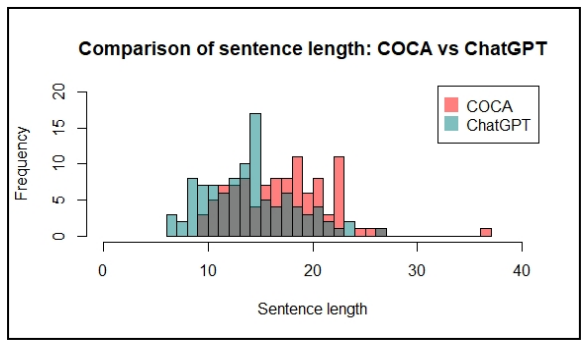

Fig. 4: Histograms presenting sentence lengths in the original dataset (COCA) and in the dataset processed by ChatGPT.

The original sentences have a mean length of 17.6 words, while the ones modified with Grammarly are 2.7 words shorter (14.9 words). The mean length of sentences processed by ChatGPT is 14.4, so 3.2 words shorter.

Therefore, one can conclude that both Grammarly and ChatGPT reduced the sentence length quite substantially. The changes they applied were more intricate than just removing *in order* in front of the *to*-infinitive, because if that was the case, there would only be a two-word difference for Grammarly (as it suggest to remove all the instances of it) and a slightly lower, but very similar difference for ChatGPT (since it removed 95% the instances). Interestingly, Grammarly shortened all 100 sentences in the sample. In the most extreme case, the reduction was by as much as sixteen words (from 37 to 21). On the other hand, ChatGPT also made a few sentences longer than they were at first. For example, one sentence enhanced by ChatGPT was seven words longer; however, even in such a case, the construction *in order to* was removed. Sentences (1)-(3) contain an example of changes that were made to the original sentences, like (1), by Grammarly, see (2), and ChatGPT, see (3).

(1) So we realized in order to even buy materials, we would need a warehouse. (COCA original, 14 words)

(2) So we realized that even buying materials would need a warehouse. (Grammarly, 11 words)

(3) We realized we needed a warehouse to purchase materials. (ChatGPT, 9 words)

Table 3 contains the mean sentence length values, together with the readability scores calculated for each set of sentences.

|  | Sentences from COCA | Sentences processed with Grammarly | Sentences processed with ChatGPT |
|---|---|---|---|
| Mean sentence length | 17.6 words | 14.9 words | 14.4 words |
| Flesh reading-ease test | 61.6 standard / average | 62.4 standard / average | 54.7 fairly difficult to read |

Table 3: Mean sentence length and readability scores for the three sets of sentences.

The datasets differ also with regard to the Flesh reading-ease test scores. The original sentences from COCA (Corpus of Contemporary American English) have a readability score of 61.6 on the Flesch reading-ease test. This indicates that they are moderately easy to read and understand. The sentences were processed by Grammarly have a slightly higher readability score of 62.4. This suggests that Grammarly has helped to make the sentences slightly easier to read and understand. The sentences that were processed by ChatGPT have a visibly lower readability score (of 54.7). This indicates that they are more difficult to read and understand than the original sentences from COCA or the sentences processed by Grammarly.

Overall, these scores suggest that both Grammarly and ChatGPT can have an impact on the readability of written text, but that the effects may vary depending on the specific tool or model being used. It's worth noting that ChatGPT is a language model designed to generate natural language text, rather than a tool specifically designed to improve readability, and since our request was "to make the sentences better", readability was not the main focus of ChatGPT. For Grammarly the situation might be different – as it is a writing assistant, one of its default functions is to facilitate understanding, thus, not so surprisingly, it makes the texts clearer and easier to understand.

## 4 Discussion and conclusions

The AI-based technologies tested in the present work represent different types of tools. Grammarly is a writing assistant, whereas ChatGPT is a chatbot generating new content. Despite these differences, the changes made to the test sentences by those two technologies point in the same direction: i) they seem to prefer less wordy and more concise content; ii) they suggest to remove 95%-100% of the instances of the purpose subordinator *in order to*, which can mean that more formal and elaborate phrases are dispreferred. The removal of *in order to* seems to appear across the board, regardless of the length of the original sentence. Tailoring the suggestion based on the original sentence length or level of formality was not observed: it looks like the goal of the algorithm is always to reduce wordiness. This is in line with the observation of the decrease in the mean sentence length and the fact that the modifications have not improved the readability. Perhaps, algorithms aiming at optimization of readability measures would not restrict the use of constructions such as *in order to* so severely? In other words, metrics, such as the mean sentence length or various readability measures, help interpret the results obtained from corpus analysis, hinting towards reasons behind the observed trends.

While the construction looked at in the present work – *in order to* – is presenting a decline in the frequency of use in langauge corpora such as COHA, this purpose subordinator, without doubt, still constitutes part of the core grammar of the English language. Now, the fact that AI-powered technologies prefer shorter phrases to longer and more elaborate ones goes in line with the changes happening in language, described in the Introduction. What we observe, very much looks like the output of the tech products is mirroring the ongoing language change.

However, we could go one step further and hypothesise that if Grammarly, ChatGPT, and similar AI-powered programmes with the same principles become sufficiently popular, they might become a factor influencing the English language – its syntactic usage, its structure and vocabulary.

Similarly, while the trend towards the shortening of sentence length has been present in the language for the last three to four hundred years, it seems to have accelerated since the beginning of the twentieth century. Given the output of Grammarly and ChatGPT and their strong tendency to favour shorter sentences and avoid wordiness in favour of conciseness, AI-powered language technologies may furthermore contribute to the shortening of sentence length in English.

It is easy to assume that if enough people use Grammarly (even in its most basic version, without tone and formality adjustment) in their daily life and simply accept all, or almost all suggestions given to them by the algorithms, be it for more "clarity", "less wordiness", "better tone" or anything else, the texts will very likely become more concise, less wordy, and the longer phrases are deemed to decrease in their frequency of use at a much higher pace than before. The change seems to already be on its way. As of March 2023, Grammarly offers a browser extension for Chrome, Safari, Firefox, and Edge. According to Grammarly's website[11], "the extension works on popular websites and can help you check your text whenever you write online." For developers, Grammarly Text Editor SDK is offered, which "can bring real-time writing support to your app by adding just a few lines of code[12]." So, with e.g. Grammarly, or other software, embedded on the websites we use daily, be it on social media, blogs, email, and wherever else, the AI-based technologies are influencing our language to a significant extent, even if we are not fully aware of it.

Also regarding the generation of new content, the present study shows that ChatGPT vastly removes *in order to* when asked to "write sentences in a better way." We can then extrapolate this observation and assume that, while creating new content, it will avoid this construction for similar reasons.

---

[11] See Grammarly's oficial website - https://support.grammarly.com/hc/en-us/articles/115000091592-Grammarly-s-browser-extension-user-guide, accessed on May 31, 2023.

[12] See Grammarly for Developers website - https://developer.grammarly.com/, accessed on May 31, 2023.

To conclude, the observations made and presented in this paper hint at the possibility that AI-powered language technologies have the potential to influence the language and accelerate language change by fostering the trends which are already present in the language. From the case studies analysed here it is clear that both tools used in the present work influence the language produced by its users. Also, it seems clear, that as of 2023, we can detect certain patterns characterizing the language of Grammarly-enhanced texts, such as, for example an absolute avoidance of phrases such as *in order to*.

Naturally, further research is needed to look at other phenomena happening in the language and the ways technologies such as Grammarly and ChatGPT are dealing with them. In addition, since, in a very unlikely scenario, the observed effects could occur due to a software bug, a similar investigation shall be repeated in the future with new releases of the tools. Similarly, further research is needed to explore the presence of possible patterns and correlations which may be present in Grammarly's algorithms and which may influence the output both on sentence- and on text-level. Another aspect, which is currently looked at by the Author of the present work, is the interplay between e.g. the genre of the sentence and the influence of the target audience setting in Grammarly on the enhancements proposed.

Even though AI-powered technologies making use of natural language processing are not omnipresent yet, with the rapid increase in popularity of tools such as Grammarly, Wordtune, ProWritingAid, or even chatbots like ChatGPT, we might be witnessing the rise of a new higher-order process influencing the language.

## *Acknowledgements*

The financial support by the University of Gdańsk (grant "UGrants-first," decision no. 1220.6010.192.2023) is acknowledged.

I want to sincerely thank the three anonymous Reviewers for their thorough reviews, which helped to make this paper better than it was at first.

## *References*

Akimoto, M., Kytö, M., Scahill, J., and Tanabe, H. 2010. *Language change and variation from Old English to Late Modern English: a festschrift for Minoji Akimoto*: Vol. 114. Peter Lang

Barrot, J. S. 2020. Integrating Technology into ESL/EFL Writing through Grammarly. *RELC Journal* 3368822096663.

Biber, D. and Conrad, S. 2009. *Register, Genre, and Style*. Cambridge: Cambridge University Press.

Crystal, D. 2004. *Making Sense of Grammar*. Pearson Longman.

Davies, M. 2008-. *The Corpus of Contemporary American English* (COCA): 520 million words, 1990-present. Available online at http://corpus.byu.edu/coca/

Davies, M. 2010-. *The Corpus of Historical American English* (COHA): 400 million words, 1810- 2009. Available online at https://corpus.byu.edu/coha/

Doshi, R. H., Bajaj, S. S., and Krumholz, H. M. 2023. ChatGPT: Temptations of Progress. *American Journal of Bioethics*, 1–3.

Durkin, P. 2014. *Borrowed Words: A History of Loanwords in English*. Oxford: Oxford University Press.

Fahnestock, J. 2011. *Rhetorical Style: The Uses Of Language In Persuasion*. New York: Oxford University Press.

Filipan-Žignić, B., Legac, V., and Sobo, K. 2016. The influence of the language of new media on the literacy of young people in their school assignments and in leisure. *Linguistics Beyond and Within* (LingBaW), 2, 77–96.

Fowler, H. W. and Butterfield, J. (ed.). 2015. *Fowler's Dictionary of Modern English Usage. 4th ed*. Oxford: Oxford University Press.

Fries, U. 2010. Sentence Length, Sentence Complexity and the Noun Phrase in 18th-Century News Publications. In Merja Kytö, John Scahill and Harumi Tanabe (eds.), *Language Change and Variation from Old English to Late Modern English* [Linguistic Insights 114], 21–33. Bern: Peter Lang.

Geertsema, S., C Hyman, C., and Van Deventer, C. 2011. Short message service (SMS) language and written language skills: educators' perspectives. *South African Journal of Education*, 31(4), 475–487.

Gordijn, B., and Have, H. T. 2023. ChatGPT: evolution or revolution? *Medicine, Health Care, and Philosophy*, 26(1), 1–2.

Gross, A. G., Harmon J. E. and Reidy M. S. 2002. *Communicating science: the scientific article from the 17th century to the present*. Oxford: Oxford University Press.

Hames, T. and Rae, N. C. 1996. *Governing America: History, Culture, Institutions, Organisation, Policy*. Manchester University Press.

Hickey, R. 2012. Internally and externally motivated language change. In *The Handbook of Historical Sociolinguistics*, Juan Manuel Hernández-Compoy and Juan Camilo Conde-Silvestre (eds.), 401–421. Malden. MA: Wiley-Blackwell.

Hilpert, M. 2013. *Constructional change in English: developments in allomorphy, word formation, and syntax*. Cambridge: Cambridge University Press.

Hiltunen, R. 1983. *The decline of the prefixes and the beginnings of the English phrasal verb. The evidence from some Old and Early Middle English texts*. Turku: Turun Yliopisto.

Hundt, M. 2014. The demise of the being to V construction. *Transactions of the Philological Society*, 112(2). 167–187.

Jin, T., and Lu, X. 2018. A Data-Driven Approach to Text Adaptation in Teaching Material Preparation: Design, Implementation, and Teacher Professional Development. *TESOL Quarterly*, 52(2): 457-467.

Kempf, L. 2021. German so-relatives: Lost in grammatical, typological, and sociolinguistic change. In Svenja Kranich and Tine Breban (eds.), *Lost in change: Causes and processes in the loss of grammatical elements and constructions*, 291–332. Amsterdam and Philadelphia: John Benjamins.

Klare, G. R. 1975. Readability and the behaviour of readers. In Shelley A., Harrison, Lawrence M. Stolurow, *Improving Instructional Productivity in Higher Education*, 235–240. New Jersey: Educational Technology Publications Englewood Cliffs.

Lewis, E. H. 1894. The History of the English Paragraph. The University of Chicago Press.

Looi, M-K. 2023. Sixty seconds on . . . ChatGPT. BMJ (Online), 380, 205–205.

Los, B. 2005. *The Rise of the To-Infinitive*. Oxford: Oxford University Press.

Mair, C. 2006. *Twentieth-century English: History, variation and standardization*. Cambridge: Cambridge University Press.

Mehrabi Boshrabadi A., and Bataghva Sarabi, S. 2016. Cyber-communic@tion etiquette: The interplay between social distance, gender and discursive features of student-faculty email interactions. *Interactive Technology and Smart Education*, 13(2), 86–106.

Naderi, B., Mohtaj, S., Ensikat, K., and Möller, S. 2019. Subjective Assessment of Text Complexity: A Dataset for German Language. ArXiv abs/1904.07733.

O'Neill, R. and Russell, A. 2019. Stop! Grammar time : University students' perceptions of the automated feedback program Grammarly. *Australasian Journal of Educational Technology*, 35(1), 42–56

Petré, P. 2010. The functions of weorðan and its loss in the past tense in Old and Middle English. *English Language and Linguistics*, (14), 457–484.

Rudnicka, K. 2018. Variation of sentence length across time and genre: influence on the syntactic usage in English. In *Diachronic Corpora, Genre, and Language Change*, Richard Jason Whitt (ed.), 219 – 240. Amsterdam and Philadelphia: John Benjamins.

Rudnicka, K. 2019. *The Statistics of Obsolescence: Purpose Subordinators in Late Modern English*. NIHIN Studies. Freiburg: Rombach.

Rudnicka, K. 2021a. The "negative end" of change in grammar: terminology, concepts and causes. *Linguistics Vanguard* 7(1), 20200091.

Rudnicka, K. 2021b. In order that – a data-driven study of symptoms and causes ofobsolescence. *Linguistics Vanguard* 7(1), 20200092.

Rudnicka, K. 2021c. So-adj-a construction as a case of obsolescence in progress. In *Lost in Change. Causes and Processes in the Loss of Grammatical Elements and Constructions* [Studies in Language Companion Series 218], Svenja Kranich and Tine Breban (eds.), 51 – 73. Amsterdam: John Benjamins.

Schmidtke-Bode, K. 2009. *A Typology of Purpose Clauses*. Amsterdam and Philadelphia: John Benjamins.

Tagg, C. 2015. *Exploring Digital Communication: Language in Action*. Routledge.

Van Dis, E., Bollen, J., Zuidema, W., van Rooij, R., and Bockting, C. L. 2023. ChatGPT: five priorities for research. *Nature* (London), 614 (7947), 224–226.

Westin, I. 2002. *Language Change in English Newspaper Editorials*. Amsterdam: Rodopi.